\documentclass[journal]{IEEEtran}

\usepackage{amsmath,graphicx}

\usepackage{amsmath}

\usepackage{enumerate}

\usepackage{algorithmic}

\usepackage{amssymb}
\usepackage{amsthm}

\usepackage{subeqnarray}

\usepackage[ruled,vlined]{algorithm2e}
\usepackage{epsfig}

\usepackage[lofdepth,lotdepth]{subfig}
\usepackage{float}
\usepackage{subfig}

\begin{document}

\title{Deep Learning with unsupervised data labeling for weeds detection on UAV images}
%
%
%

\author{
	\IEEEauthorblockN{M.Dian. Bah\IEEEauthorrefmark{1}, Adel Hafiane\IEEEauthorrefmark{2}, Rapha\"el Canals\IEEEauthorrefmark{1}}\\
	\IEEEauthorblockA{\IEEEauthorrefmark{1}University of Orl\'{e}ans, PRISME EA 4229,  Orl\'{e}ans F45072, France} \\
	\IEEEauthorblockA{\IEEEauthorrefmark{2}INSA Centre Val de Loire, PRISME EA 4229, Bourges F-18022, France}
	
}

\maketitle

\begin{abstract}
	
	In modern agriculture, usually weeds control consists in spraying herbicides all over the agricultural field.  This practice involves significant waste and cost of herbicide for farmers and environmental pollution. One way to reduce the cost and environmental impact is to allocate the right doses of herbicide at the right place and at the right time (Precision Agriculture). Nowadays, Unmanned Aerial Vehicle (UAV) is becoming an interesting acquisition system for weeds localization and management due to its ability to obtain the images of the entire agricultural field with a very high spatial resolution and at low cost. Despite the important advances in UAV acquisition systems, automatic weeds detection remains a challenging problem because of its strong similarity with the crops. Recently Deep Learning approach  has shown impressive results in different complex classification problem. However, this approach needs a certain amount of training data but, creating large agricultural datasets with pixel-level annotations by expert is an extremely time consuming task.  In this paper, we propose a novel   fully automatic learning method using Convolutional Neuronal Networks (CNNs) with unsupervised training dataset collection for weeds detection from UAV images. The proposed method consists in three main phases. First we automatically detect the crop lines and using them to identify the interline weeds. In the second phase, interline weeds are used to constitute the training dataset. Finally, we performed CNNs on this dataset to build a model able to detect the crop and weeds in the images. The results obtained are comparable to the traditional supervised training data labeling.  The accuracy gaps  are 1.5\% in the spinach field and 6\% in the bean  field.
	
\end{abstract}


\section{Introduction}
\label{intro}

Currently,  losses due to pests, diseases and weeds can reach 40\% of global crop yields each year and this
percentage is expected to increase significantly in the coming years \cite{EuropeanCropProtection2017}. The usual weeds control practices consist in spraying herbicides all over the agricultural field. Those practices involve significant wastes and costs of herbicides for farmers and environmental pollution \cite{Oerke2006}.  In order to reduce the amount of chemicals while continuing to increase productivity, the concept of  precision agriculture was introduced \cite{Pierce1999,McBratney2005}. Precision agriculture is defined as the application of technology for the purpose of improving crop performance and environmental quality \cite{Pierce1999}. The main goal of precision agriculture is to allocate the right doses of input at the right place and at the right time. Weeds detection and characterization represent one of the major challenges of the precision agriculture. 

Unmanned Aerial Vehicle (UAV) is fast becoming an interesting vision based acquisition system, since it enables rapid acquisition of the entire crop area with a very high spatial resolution and at  low cost \cite{Torres-Sanchez2015,Zhang2012}.  Despite the important advances in UAV acquisition systems, the automatic detection of weeds remains a challenging problem. In recent years, deep learning techniques have shown a dramatic improvement for  many computer vision tasks, but still not widely used in agriculture domain. Indeed recent development showed the importance of these techniques for weeds detection \cite{DosSantosFerreira2017,Bah2018}. However, the huge quantities of the data required in the learning phase, have accentuated the problem of the manual annotation of these datasets. The same problem rises in agriculture data, where  labeling plants in a field image is very time consuming. So far, very little attention have been payed to the unsupervised annotation of the data to train the deep learning models, particularly for agriculture.   

In this paper, we propose a new fully automatic learning method using Convolutional Neuronal Networks (CNNs) with unsupervised training set construction for weeds detection on UAV images. This method is performed in three main phases. First we automatically detect the crop lines and using them to identify the interline weeds. In the second phase interline weeds are used to constitute our training dataset. Finally, we performed CNNs on this database to build a model able to detect the crop and weeds in the images.


This paper is divided into five parts. In the section 2 we discuss the related work. Section 3  presents the proposed method. In section 4 we comment and discuss the experimental results obtained. Section 5 concludes the paper.

\section{Related work}

In literature, several approaches have been used to detect weeds with different acquisition systems.  The main approach for weeds detection is to extract vegetation from the image using a segmentation and then  discriminate crop and weeds. Common segmentation approaches use  color and multispectral
information, to separate vegetation and background (soil and residues). Specific indices are calculated from these information to effectively segment vegetation \cite{Hamuda2016}. 

However, weeds and crop are hard to discriminate by using spectral information because of their strong similarity. Regional approaches and spatial arrangement of pixels are preferred in most cases.  In \cite{Gee2008},  Excess Green Vegetation Index (ExG) \cite{Woebbecke1995} and the Otsu's thresholding \cite{Otsu1979} have helped to remove background (soil, residues) before to perform a double  Hough transform \cite{Hough1962} in order to identify the main crop lines in perspective images. Then, to discriminate crop and weeds in the segmented image,  the authors applied a region-based segmentation method developing a blob coloring analysis. Thus any region with at least one pixel belonging to the detected lines is considered to be crop, otherwise it is weeds. Unfortunately, this technique failed to handle weeds close to crop region.  In \cite{Pena2013} an object-based image analysis (OBIA) procedure was developed on  series of UAV images  for  automatic discrimination of crop rows and weeds in maize field. For that, they segmented the UAV images  into homogeneous multi-pixel objects using the multi-scale algorithm \cite{Blaschke2010}. Thus, the large scale highlights structures of crop lines and the small scale brings out objects that lie
within crop lines. They have found that the process is strongly affected by the presence of weed plants very close or within the crop rows.

In \cite{Tang2003}, 2-D Gabor filters was applied to extract the features and ANN for broadleaf and grass weeds classification. Their results showed that joint space-frequency texture features have potential for weed classification.  In \cite{Jeon2011}, the authors rely on morphological variation and use neural network analysis to separate weeds from maize crop. Support Vector Machine (SVM) and shape features was suggested for the effective classification of crops and weeds in digital images in \cite{Ahmed2012}. On their experiment, a total of fourteen features that characterize crops and weeds in images were tested to find the optimal combination of features which provides the highest classification rate. \cite{Latha2014} suggested that  in the image, edge frequencies and veins of both the crop and the weed have different density properties (strong and weak edges) to separate crop from weed.  A semi-supervised method has been proposed in \cite{Perez-Ortiz2015a} to discriminate weeds and crop. The Ostu thresholding was applied twice on ExG. In first step, authors used segmentation to remove the background then, in the second one they create two classes  supposed to be crop and weeds. K-means clustering was used to select one hundred samples of each class for the training. SVM classifier with geometric features, spatial features, first and second-order statistics was extracted on the red, blue, green and ExG bands. The method has proven to be effective in sunflower field, but less robust in the corn field because of shade produced by corn plants.

In \cite{Bakhshipour2017}, authors  used texture features extracted from wavelet sub-images to detect and characterize four types of weeds in a sugar beet field. Neural networks have been applied as classifier. The use of wavelets proved to be  efficient for the detection of weeds even at a stage of growth of beet  greater than 6 leaves.  \cite{Bakhshipour2018} evaluate  weeds detection with support vector machine and artificial neural networks in four species of common weeds in sugar beet fields using shape features.

Recently, convolutional neural networks have emerged as a powerful approach for computer vision tasks.  CNNs  \cite{LeCun1998} progressed mostly through the success of this method in  ImageNet Large Scale Vision Recognition Challenge 2012 (ILSCVR12) and the creation of AlexNet network in 2012 which  showed that a large, deep convolutional neural network is capable of achieving record-breaking results on a highly challenging dataset using purely supervised training \cite{Krizhevsky2012}. Nowadays  deep learning is applied in several domains  to help solve many big data problems such as computer vision, speech recognition, and natural language processing.  In agriculture domain,  CNNs  are applied to classify patches of water hyacinth, serrated tussock and tropical soda apple in \cite{Hung2014}. \cite{Mortensen2016} used CNNs  for semantic segmentation in the context of mixed crops on images of an oil radish plot trial with barley, grass, weed, stump and soil.  \cite{Milioto2017} provide accurate weeds classification in real sugar beet fields with mobile agricultural robots.   \cite{DosSantosFerreira2017} applied AlexNet  for the detection of weeds in soybean crops. In \cite{Bah2018} AlexNet is applied for weeds detection in different crop fields such as the beet, spinach and bean in  UAV imagery.

The main common point between the supervised machine learning  algorithms is the need of training data. For a good optimisation of deep learning models it is necessary to have a certain amount of labeled data. But  as mentioned before creating large agricultural datasets with pixel-level annotations is an extremely time consuming task. 

Little attempts have been made to develop fully automatic system for training and identification of weeds in agricultural fields. In a Recent work,  \cite{DiCicco2017} suggest the use of synthetic training datasets. However, this technique  requires a precise modeling in terms of texture, 3D models and light conditions.

To the best of our knowledge there is no work for automatic labeling of weeds using UAV imagery system.

\section{Proposed Method}

\label{method}

\begin{figure*}[h]
	\centering
	
	\includegraphics[width=15cm]{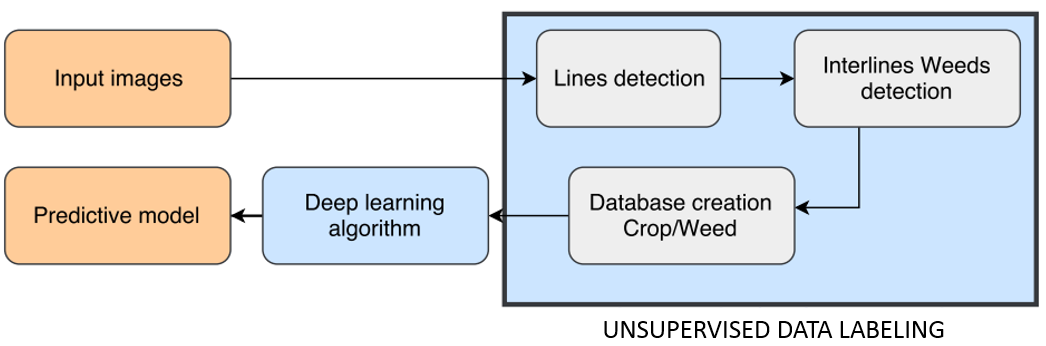}
	\caption{ Flowchart of the proposed method}
	\label{fig:flowchart}
\end{figure*}

In modern agriculture, most of crops are grown in regular rows separated by a defined space that depends on type of the crop.  Generally, plants that grow out of the rows are considered as weeds commonly referred as inter-line weeds. Several studies have used this assumption to locate weeds using the geometric properties of the rows \cite{Jones2009}. The main advantage of such technique is that it is unsupervised and does not depend on the training data. Indeed, based on this hypothesis. Intra and inter line vegetation are then used to constitute our training database which is categorized into two classes crop and weed.  Thereafter, we performed CNNs on this database to build a model able to detect the crop and weeds in the images.  The flowchart  (Fig \ref{fig:flowchart}) depicts the main steps of the proposed method. Next sections describes in details each step.

\subsection{Crop lines detection}

A crop row can be defined as a composition of several parallel lines. The aim is to detect the main line of each crop row. For that purpose we have used a Hough transform to highlight the alignments of the pixels. In Hough space there is one cell by line which involve an aggregation of cells by crop row. The main lines in Hough space correspond to the cells which contains the maximum of vote (peak) on each aggregation. 
Before starting any line detection procedure, generally pre-processing is required to remove undesirable perturbations such as shadows, soil or stones. Here we have used the ExG   (Eq. \ref{exg}) with the Otsu adaptive thresholding  to discriminate between vegetation and background.

\begin{equation}
	\label{exg}
	ExG=2 g - r - b  		
\end{equation}
where r, g and b are the normalized RGB coordinates.

\begin{figure*}
	\centering
	\includegraphics[height=9cm, width=15cm]{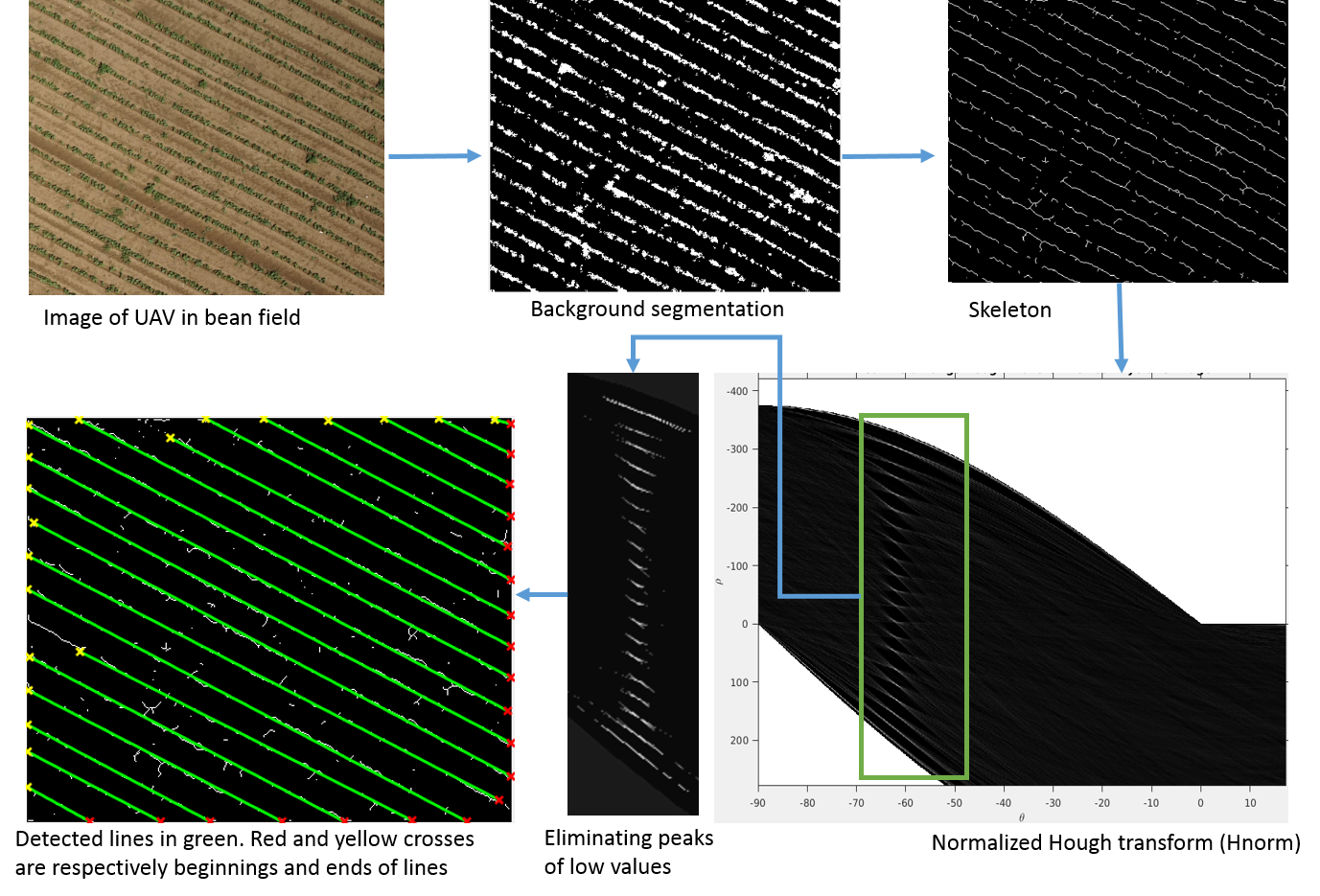}
	\caption{ Flowchart of crop lines detection method.  }
	\label{fig:Lines}
	
\end{figure*}

Hough transform is one of the most widely used methods for lines detection and it is often integrated in  tools for guiding agricultural machines because of  its robustness and  ability to adjust discontinuous lines caused by missing crop plants in the row or  poor germination  \cite{Montalvo2012}. 
Usually, for  crop lines detection,  Hough transform  is directly applied to the segmented image. This procedure is computational expensive and depends  on the density of the vegetation in crop rows and there is also a  risk of the lines over-detection. 
We have addressed this problem by using the skeleton of each row instead of the binary region, this approach  has shown  better performances in (Bah et al. 2017). The skeleton provided a good overall representation of the structure of the field, namely orientations and periodicity.  Indeed, the Hough transform  $H(\theta,\rho)$ is computed  on the skeleton with a $\theta$ resolution equal to 0.1$^{\circ}$ letting $\theta$ take values in the range of  $]-90^{\circ};90^{\circ}]$ and $\rho$ resolution equal to 1.  Thanks to a histogram of the skeletons directions,  the most represented angle is chosen as the main orientation $\theta_{lines}$ of  crop lines.  $H(\theta,\rho)$ has been normalized  $H_{norm}$($\theta,\rho$) in order to give the same weight to all the crop lines, especially the short ones close to the borders of the image \cite{Gee2008}. $H_{norm}$($\theta,\rho$)  is defined as the ratio between the accumulator of the vegetation image and the accumulator of a totally white image of the same size $H_{ones}$($\theta,\rho$). To disregard the small lines created by aggregation of weeds in inter-row a threshold of 0.1 was applied to the normalized Hough transform. Moreover  in modern agriculture crops are usually sown in parallel lines with the same interline distance that is the main  peaks corresponding to the crop lines are aligned around an angle in the Hough space with same  gaps. Unfortunately, because of the realities in the agricultural field  the lines are not perfectly parallel, thus the peaks in the Hough space have close but different angles and the interline distance is not constant. In order to not skip any crop line during the detection, all the lines which have in Hough space a peak whose angle compared to the overall orientation ($\theta_{lines}$)  of the lines does not exceed 20$^ {\circ} $ are retained. Fig. \ref{fig:Lines} presents the flowchart of the lines detection method. However, to avoid detecting more than one peak in an aggregation (reduce over-detection), every time  that we identify a peak of a crop row in $H_{norm}$($\theta,\rho$),  we identify the corresponding skeleton, and then we delete the votes of this skeleton in $H_{norm}$($\theta,\rho$) before continuing.   All the steps are  summarized in algorithm 1.

\begin{algorithm}
	
	\DontPrintSemicolon
	
	\SetKwData{Left}{left}\SetKwData{This}{this}\SetKwData{Up}{up}
	
	\SetKwFunction{Union}{Union}\SetKwFunction{FindCompress}{FindCompress}
	
	\SetKwInOut{Input}{input}\SetKwInOut{Output}{output}
	
	\Input{skeletons}
	
	\Output{crop lines }
	
	\nl Computation of the skeletons angle \;
	
	\nl Computation of  the main orientation $\theta_{lines}$ of  the crop lines\;
	
	\nl Hough transform of the skeletons $H(\theta,\rho)$\;
	
	\nl $H_{norm}$($\theta,\rho$)=$H(\theta,\rho) $/Hones($\theta,\rho$)\;

	\nl \While{maximum of $H_{norm}$($\theta,\rho$) $>$ 0.1}{
		
		\nl Computation of the maximum of $H_{norm}$($\theta,\rho$) and the corresponding angle $\theta_m$ \;

		\nl Recovery of the line corresponding to the maximum ($Line_{skeleton}$)\;
		
		\nl Computation of the normalized  Hough transform  ($H_{temp}$($\theta,\rho$)) of the $Line_{skeleton}$   \;
		
		\nl	$H_{norm}$($\theta,\rho$)=$H_{norm}$($\theta,\rho$) $- H_{temp}(\theta,\rho)$\;
		
		\nl \If { $\theta_m > \theta_{lines} -20^ {\circ} $ and $\theta_m < \theta_{lines} +20^ {\circ} $}{
			
			\nl The detected line is a crop line\;	 
			
		}  
		
	}
	
	\caption{Crop lines detection. }
	
\end{algorithm}

\begin{figure*}[h]
	\centering
	\subfloat[]{\label{fig:L-a2} \includegraphics[height=6cm, width=7cm]{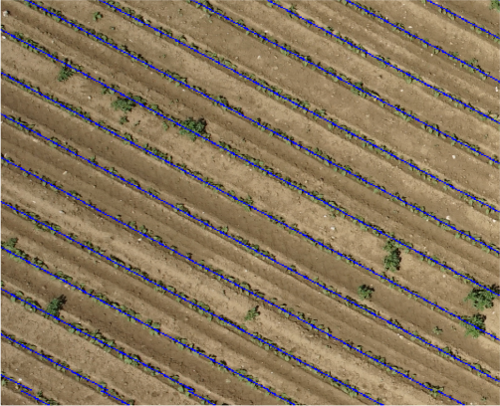}}
	\hspace{1pt}
	\subfloat[]{\label{fig:L-b2} \includegraphics[height=6cm, width=7cm]{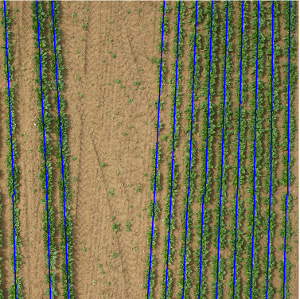}}
	
	\caption{ From left to right lines detection in bean (a) and spinach (b) fields. Detected lines are in blue. In the spinach field interline distance and the crop rows  orientation are not regular. The detected lines are mainly in the center of the crop rows. }
	\label{fig:Interline_weeds}
\end{figure*}


\begin{figure*}[h]
	\centering	
	
	\hspace{1pt}
	\subfloat[]{\label{fig:sup-a}\includegraphics[height=6cm, width=5cm]{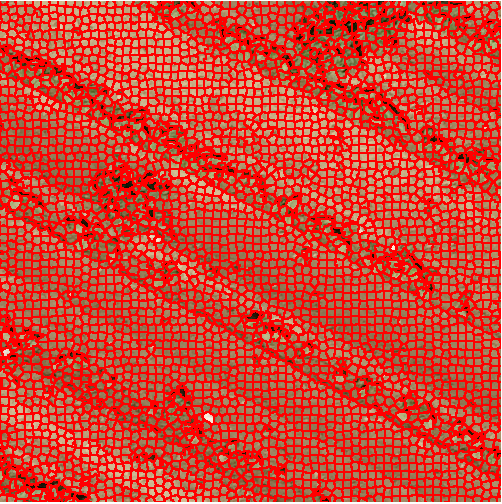}}
	\hspace{1pt}
	\subfloat[]{\label{fig:sup-b}\includegraphics[height=6cm, width=5cm]{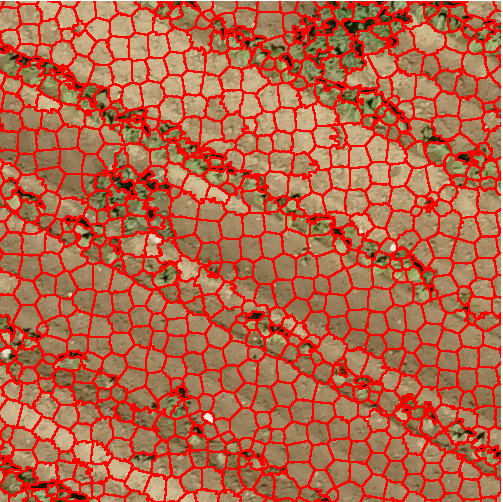}}
	\hspace{1pt}
	\subfloat[]{\label{fig:sup-c}\includegraphics[height=6cm, width=5cm]{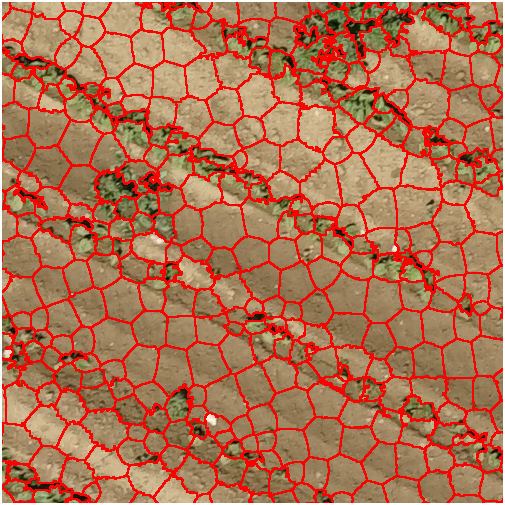}}
	\caption{Examples of superpixels computed on images of dimensions N=7360$\times$4912. From left to right, the image  is segmented with a number of superpixels equal to  0.5\% $\times$ N, 0.1\%$\times$N and 0.01\% $\times$N respectively.}
	\label{fig:superPixels}
\end{figure*}
\subsection{Unsupervised training data labeling}
\label{Unsup_data}
The unsupervised training dataset annotation is based on the detected lines obtained in previous section. 
According to hypothesis that the lines detected are mainly at the center of the crop rows  (Fig. {\ref{fig:Interline_weeds}) we performed a  mask  to delimit the crop rows. Hence, the vegetation overlapped by the mask  correspond to the crop. This mask is obtained from the intersection of superpixels formed by the simple linear iterative clustering (SLIC) algorithm \cite{Achanta2011} and the detected lines. SLIC is chosen since it is simple and efficient in terms of results quality and computation time. It is an adaptation of k-means for superpixels generation with a control on  size and compactness of superpixels. SLIC creates a local grouping of pixels based on their spectral values defined by the values of the CIELAB color space and their spatial proximity. A higher value of compactness makes superpixels more regularly shaped. A lower value makes superpixels adhere to boundaries better, making them irregularly shaped.
	Since here the goal is to create a mask around the detected crop lines able to delimit the crop rows we have chosen a compactness of 20 because we found it was less sensitive to the variation of color caused by the effect of light and shadow. Fig. \ref{fig:superPixels} shows examples of images segmented with different size of superpixels.

	Once the crop is identified, next step consists in detection of the interlines weeds. 
	Interline weed is plant which grows up in the interline crop. To detect weeds that lie in inter-row we performed a blob coloring algorithm. Hence any region that does not intersect with the crop mask is regarded as weed. Besides, vegetation pixels which do not belong to the crop mask neither to the interlines weeds are attributed to the potential weeds. Fig. \ref{fig:samples_weeds}  shows the mask of crop, interline weeds and potential weeds.
	To construct the training dataset, we extracted patches from the original images using positions of the detected inter-row weeds and crops.  For weeds samples we performed  bounding boxes on each segmented intra-row weed. For the crop samples, sliding window has been applied on the input image using  positions relative to the segmented crop lines.  Thus, for a given position  of the window  if it intersects the  binary mask  and there is no inter-lines weeds pixels  we attribute it to the crop class.
	Generally, the crop class has much more samples than the weed. In the case that we have less interline weeds  samples and in the same time we have a wide potential weeds  as in Fig.\ref{fig:samples_weeds} we propose to collect  samples from the potential weeds. Hence, the window which contains only potential weeds is labeled as weeds. Windows which contain crop and potential weeds, where we have more potential weeds than crop are not retained.  
	
	\begin{figure}[h]
		\centering
		
		\includegraphics[scale=0.25]{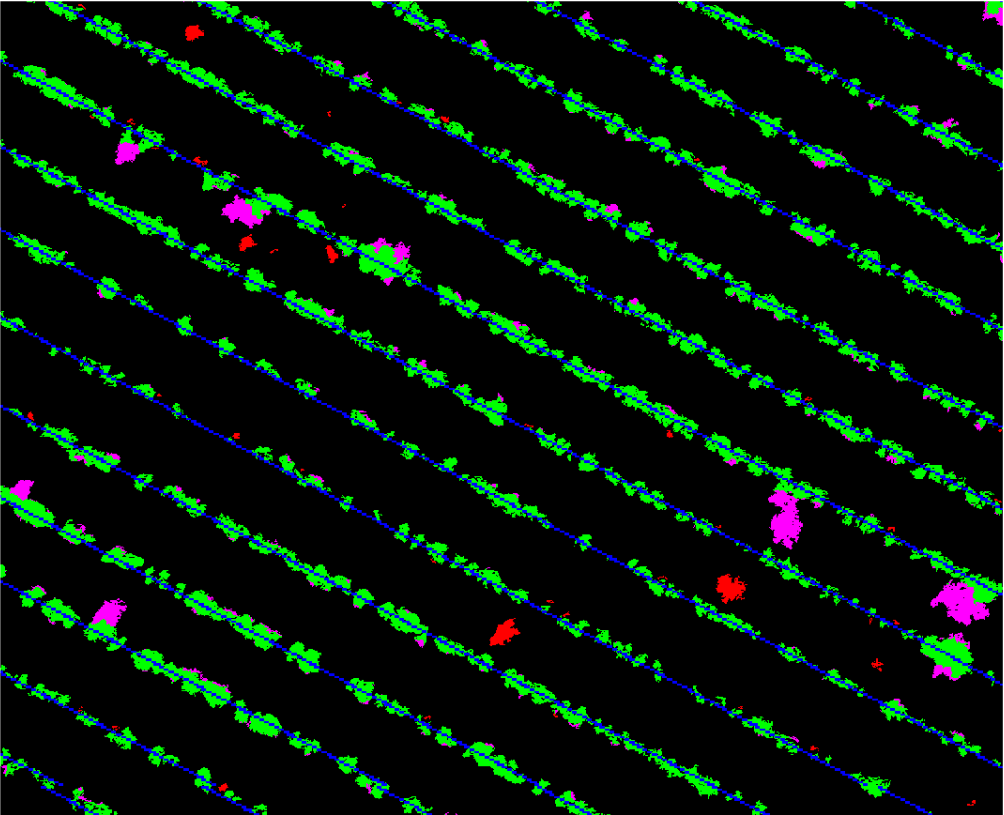}
		\caption{Detection of interline weeds (red) after lines detection (blue) in a bean image. The mask of crop is represented in green and the potential weeds in magenta.}
		\label{fig:samples_weeds}
	\end{figure}

	\subsection{Crop/Weeds classification using Convolutional Neural Networks}
	CNNs are a part of Deep learning approach, they showed impressive performances in many computer vision tasks \cite{Kamilaris2018}.  CNNs are  made up of two types of layers, the convolutional layers which  extract different characteristics of images and   the fully connected layers based on multilayer perceptron.  The number of convolutional layers  depends on the classification task and also  the number  and the size of the training data.
	
	In this work we used a Residual Network (ResNet), this network architecture was introduced in 2015 \cite{He2016}.  It won the ImageNet Large Scale Vision Recognition Challenge 2015 with 152 layers.  However, according to the size of data we used the ResNet with 18 layers (ResNet18) described in \cite{He2016} because it has shown a better result than AlexNet and VGG13 \cite{Simonyan2015} in the ImageNet challenge. Due to  abundant categories and significant number of images in ImageNet, studies revealed the performance of transferability of networks trained with ImageNet dataset. Thus we performed fine tuning to train the networks  in our data. Fine-tuning means that we start with the learned features on the ImageNet dataset, we truncate the last layer (softmax layer) of the pre-trained network and replace it with  new softmax layer that are relevant to our own problem. Here the thousand categories of ImageNet have been replaced by two categories (crop and weeds).

	\section{Experiments and results}
	
	\begin{figure*}[h]
		\centering
		\subfloat[]{\label{fig:L-a3}\includegraphics[height=8cm, width=8cm]{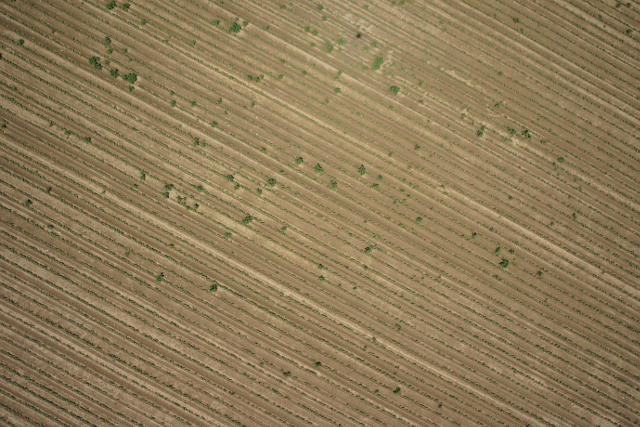}}
		\hspace{1pt}
		\subfloat[]{\label{fig:L-b3}\includegraphics[height=8cm, width=8cm]{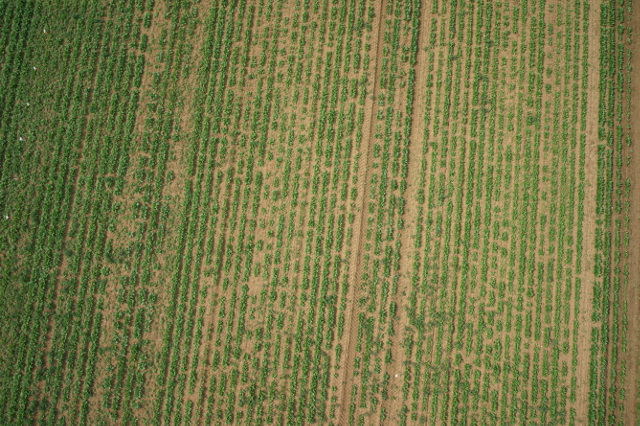}}
		
		\caption{Example of images taken in the bean (a) and spinach fields (b). The bean field has less interline weeds and  is predominately composed of  potential weeds. The inter-row distance is stable and the plant is sparse compared to the spinach field which presents a dense vegetation in the crop rows and irregular inter-row distance.  Spinach field has more interlines weeds and it has  few potential weeds.}
		\label{fig:Field}
	\end{figure*}

	\begin{figure*}
		\centering
		\subfloat[]{\label{fig:L-a3}\includegraphics[height=3cm, width=4cm]{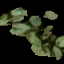}}
		\hspace{1pt}
		\subfloat[]{\label{fig:L-b3}\includegraphics[height=3cm, width=4cm]{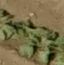}}
		\hspace{1pt}
		\subfloat[]{\label{fig:L-a3}\includegraphics[height=3cm, width=4cm]{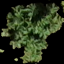}}
		\hspace{1pt}
		\subfloat[]{\label{fig:L-b3}\includegraphics[height=3cm, width=4cm]{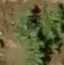}}
		\hspace{1pt}
		\subfloat[]{\label{fig:L-c3}\includegraphics[height=3cm, width=4cm]{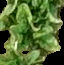}}
		\hspace{1pt}
		\subfloat[]{\label{fig:L-d3}\includegraphics[height=3cm, width=4cm]{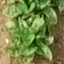}}
		\hspace{1pt}
		\subfloat[]{\label{fig:L-c3}\includegraphics[height=3cm, width=4cm]{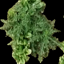}}
		\hspace{1pt}
		\subfloat[]{\label{fig:L-d3}\includegraphics[height=3cm, width=4cm]{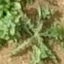}}
		\caption{Example of crop  and weed samples of size 64$\times$64 with and without background. Bean: samples of crop (a and b), samples of weed. (c and d). Spinach: samples of crop (e and f) and samples of weed (g and h). Depending on the size of the plant and the position of the window we obtain a plant or aggregation of plants per window.}
		\label{fig:samples_mask}
	\end{figure*}
	
	\begin{figure*}
		\centering
		\subfloat[]{\label{fig:L-aa} \includegraphics[height=5cm, width=8cm]{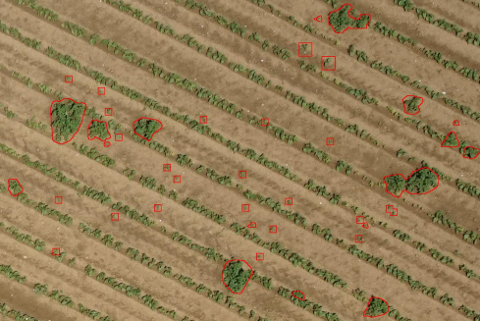}}
		\hspace{2pt}
		\subfloat[]{\label{fig:L-bb} \includegraphics[height=5cm, width=8cm]{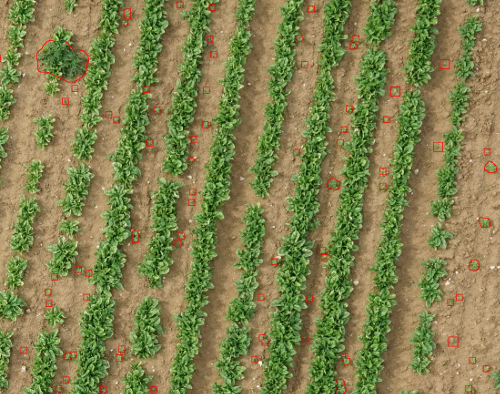}}
		
		\caption{Parts of bean field (a) and spinach field (b) with the weeds labeled manually by an expert in red. The manual labeling has taken about 2 working days.}
		\label{fig:field}
	\end{figure*}  
	
	Experiments were conducted on two different fields  of bean and spinach (Fig.\ref{fig:Field}).
	The  images are acquired by a DJI Phantom 3 Pro drone that embeds a 36 MP RGB camera at an altitude of 20 m. This acquisition system enables to obtain very high resolution images with a spatial resolution about 0.35 cm.

	To build the unsupervised training database, we selected two different parts of given field. The first one (Part1) is used to collect the  training data and the other  (Part2) for  test data collection.  
	
	To create the crop binary mask after line detection the superpixels compactness have been set to  20 and the  number of superpixels is equal to 0.1\%$\times$N, where  N=7360$\times$4912 (Fig.\ref{fig:sup-b}). In this experiments, we used a 64 by 64  window to create the weed and crop training databases. This window size  provides a good trade off between plant type and overall information. A small window is not sufficient to capture whole plant and  can lead to confuse culture and non culture, because in some conditions crop and weed leaves have the same visual characteristics. In other hand,  too large size presents a risk of having crop and weeds in the same window. 
	
	In the bean field, the weeds present are thistles and young sprouts of potato from previous sowing on the same field.  This field has few interline weeds so we have decided to include the potential weeds in weeds samples. After applying the unsupervised labeling method, the number of samples collected is 673 for weeds and 4861 for crops. Even with potential weeds the collected samples was unbalanced. To address this  problem we realized data-augmentation. Hence we have performed 2 contrast changes, a smoothing with a Gaussian filter and 3 rotations (90$^ {\circ} $, 180$^ {\circ} $, 270$^ {\circ} $).
	The strong heterogeneity in the fields can often be encountered from one part of the field to another. This heterogeneity may be a difference of soil moisture, presence of straw, etc... In order to make our models robust to the background, we mixed samples with background and no background. Samples without background were obtained by applying ExG followed by Otsu's thresholding on previously created samples (see Fig. \ref{fig:samples_mask}).
	We evaluated the performance of our  method by comparing models created by data labeled in supervised and unsupervised way.
	
	The supervised training dataset were labeled by human  experts. A mask is performed manually on the pixels of  weeds and crops. Fig. \ref{fig:field}  presents weeds  manually delineated by an expert in red.
	
	The supervised data collected were also unbalanced  so we have carried out the same data augmentation procedure performed on the unsupervised data. The total number of samples is shown in the Table \ref{SData}.
	
	The spinach field is more infected than the bean field, there are mainly thistles. In total 4303 samples of crop and 3626 samples of weed were labeled in unsupervised way. Unlike  bean field we have obtained a less unbalanced data. Therefore, the only data augmentation  applied is adding samples without background.  The same processing has been applied on the supervised data. Table \ref{SData_spinach} presents the number of samples.

	After the creation of both weed and crop classes, 80\% of samples were selected randomly for the training, and  the remaining ones were used for validation. The Tables \ref{SData} and \ref{SData_spinach} present the training and validation data performed on each field. 
	
	For finetuning we tested different values of the learning rate. The initial learning rate is set to 0.01 and updated every 200 epochs. The update is done by dividing the learning rate by factor of 10. Fig. \ref{fig:Losses} shows the evolution of the loss function during training for supervised and unsupervised datasets for spinach and bean fields. From these figures we can notice that the validation loss curves decrease during about  the first  80 epochs before to increase and to converge (behavior close to overfitting phenomenon).  This overfiting phenomenon is less emphasized in the  supervised labeled data of bean. The best models were obtained during the first learning phase with a learning rate of 0.01.

	\begin{figure*}[h]
		\centering	
		\subfloat[]{\label{fig:S-a}\includegraphics[height=5cm, width=7cm]{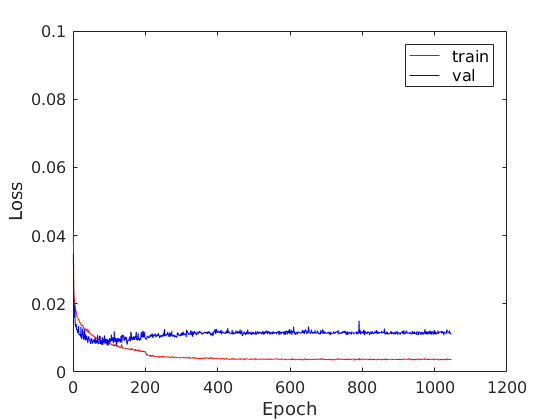}}
		\hspace{1pt}
		\subfloat[]{\label{fig:S-b}\includegraphics[height=5cm, width=7cm]{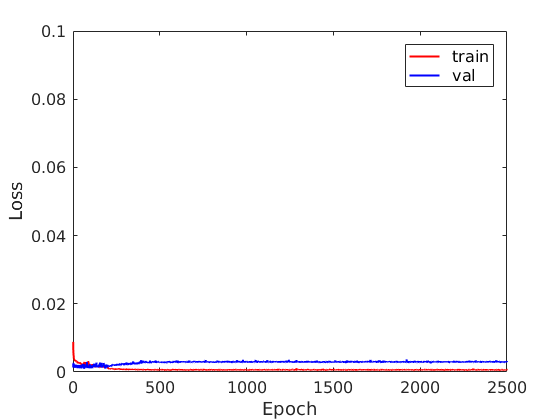}}
		\hspace{1pt}
		
		\subfloat[]{\label{fig:B-c}\includegraphics[height=5cm, width=7cm]{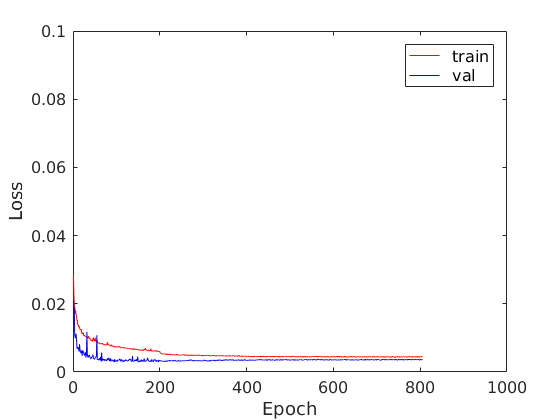}}
		\hspace{1pt}
		\subfloat[]{\label{fig:B-d}\includegraphics[height=5cm, width=7cm]{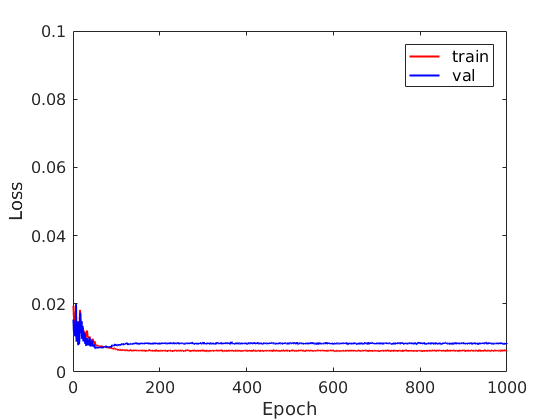}}
		
		\caption{ Evolution of the loss during training for supervised and unsupervised data in the fields of spinach and bean. The validation loss curves decrease during about the first 80 epochs before to increase and converging. First line represents the spinach field and the second one the bean field. The first and second column are respectively  the training on the the supervised and unsupervised data.}
		\label{fig:Losses}
	\end{figure*}

	\begin{figure*}[h]
		\centering	
		\subfloat[]{\label{fig:roc-a}\includegraphics[height=5cm, width=7cm]{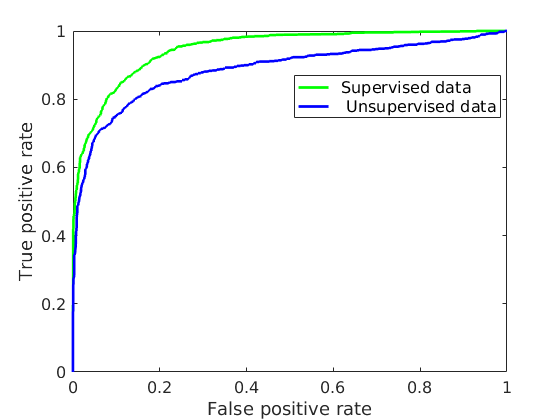}}
		\hspace{1pt} 
		\subfloat[]{\label{fig:roc-b}\includegraphics[height=5cm, width=7cm]{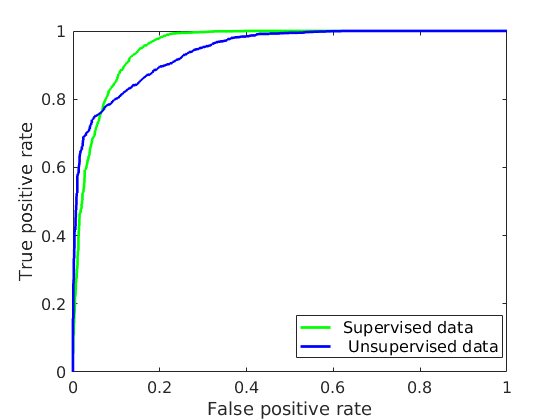}}
		\caption{ROC curves of  the test data with the unsupervised and supervised data labeling.From left to right the ROC curves computed on the bean (a)  and spinach (b) test data.}
		\label{fig:ROC}
	\end{figure*}
	
	\begin{table}
		
		\centering
		\caption{Training and validation data in the bean field.}
		
		\begin{tabular}{l c c c c }
			\textbf{Data} &\textbf{Class} &\textbf{Training}&\textbf{Validation} &\textbf{Total} \\
			\hline	
			\\
			Supervised &Crop& 17192& 11694&28886\\
			labeling&Weed&17076&9060&16136\\
			
			&Total&34868&20754&45022\\
			\hline
			\\Unsupervised &Crop&7688& 1928&9616\\
			labeling&Weed&5935&1493&7428\\
			
			&Total&13623&3421&17044\\
			\hline
			
		\end{tabular}
		\label{SData}
	\end{table}
	
	\begin{table}
		
		\centering
		\caption{Training and validation data in the spinach field.}
		
		\begin{tabular}{l c c c c }
			\textbf{Data}&\textbf{Class} &\textbf{Training}&\textbf{Validation} &\textbf{Total} \\
			\hline	
			\\
			Supervised &Crop& 11350& 2838&14188\\
			labeling&Weed&8234&2058&10292\\
			
			&Total&19584&4896&34772\\
			\hline
			\\Unsupervised&Crop& 6884& 1722&8606\\
			labeling&Weed&5800&1452&7252\\
			
			&Total&12684&3174&15858\\
			\hline
			
		\end{tabular}
		\label{SData_spinach}
	\end{table}
	
	Performance of models have been evaluated on  test ground truth data collected in Part2 by supervised way on each field, the Table \ref{TData_test} presents the samples. The performance of the classification results are illustrated with  Receiver Operating Characteristic (ROC) curves.

	\begin{table}
		
		\centering
		\caption{Number of test samples used for each field.}
		
		\begin{tabular}{l c c }
			
			\textbf{Field}&\textbf{samples of crop}&\textbf{Sample of  weed}\\
			\hline
			\\
			Bean&2139&1852\\
			Spinach&1523&1825\\
			\\
			\hline
		\end{tabular}
		\label{TData_test}
	\end{table}

	From the ROC curves (Fig.\ref{fig:ROC}) we can notice that the AUC (Area Under the Curve) are close to or greater than 90\%. Although both types of learning data provide good results and their results are comparable. On both fields we remark that positive rate of 20\%  provides a true positive rate greater than 80\%. The  differences of performance between supervised and unsupervised data labeling  are about 6\% in the bean field and about 1.5\% in the spinach field (Table \ref {AUC_resnet}). The performance gap in the bean field can be explained  by the low presence of weeds in the inter-row.
	
	\begin{table}
		
		\centering
		\caption{Results on test data with ResNet18 network.}
		
		\begin{tabular}{l c c c }
			
			\textbf{Field}&\textbf{AUC\% of  unsupervised }&\textbf{AUC\%  supervised }\\
			& \textbf{data labeling}&\textbf{data labeling}\\
			\hline
			\\
			Bean&88.73&94.84\\
			Spinach&94.34&95.70\\
			\\
			\hline	
		\end{tabular}
		\label{AUC_resnet}
	\end{table}
	
	Both fields are infested mainly by thistles, we tested the robustness of our models by exchanging the samples of weeds from the bean field with that of the spinach field. 
	
	\begin{figure*}[h]
		\centering	
		\subfloat[]{\label{fig:Exchange_weeds-a}\includegraphics[height=5cm, width=7cm]{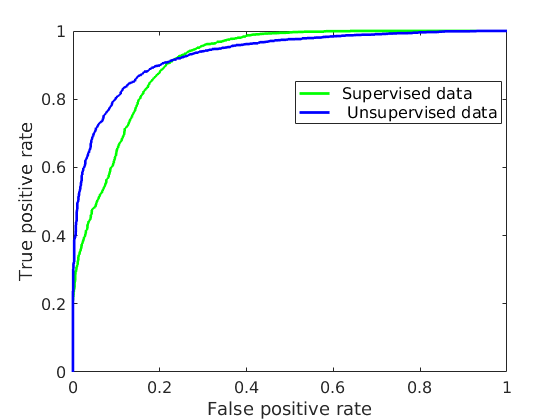}}
		\hspace{1pt}
		\subfloat[]{\label{fig:Exchange_weeds-b}\includegraphics[height=5cm, width=7cm]{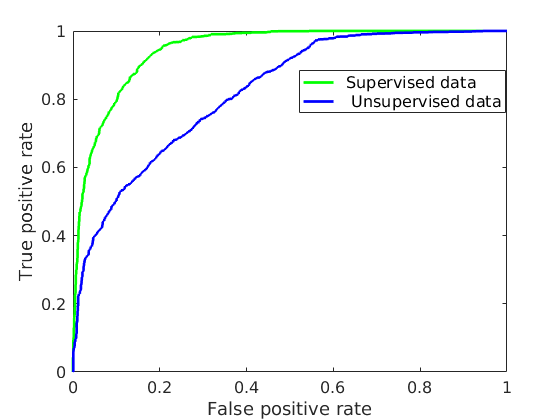}}
		
		\caption{ROC curves of  test data with  weeds data of the bean field exchanged with that of the spinach field. From left to right the ROC curves computed on the bean (a)  and spinach (b) test data. }
		\label{fig:Exchange_weeds}
	\end{figure*}
	
	In Fig. \ref{fig:Exchange_weeds}  the results obtained show that despite the small samples harvested in the bean filed, those data are suitable to the spinach field and the model created with unsupervised labeling in the spinach field is most sensitive to the presence of young potato sprouts among bean weed samples. Table \ref{AUC_resnet_robustness} shows the results on test data with weeds data of the bean field exchanged with that of the spinach field.

	\begin{table}
		
		\centering
		\caption{Results on test data with  weeds data of the bean field exchanged with that of the spinach field.}
		
		\begin{tabular}{l c c c }
			
			\textbf{Field}&\textbf{AUC\% of  unsupervised }&\textbf{AUC\%  supervised }\\
			& \textbf{data labeling}&\textbf{data labeling}\\
			
			\hline
			\\
			Bean&91.37&93.25\\
			Spinach&82.70&94.34\\
			\\
			\hline	
		\end{tabular}
		\label{AUC_resnet_robustness}
	\end{table}
	
	In addition to the classification evaluations on the patches of images, we applied an overlapping window to classify each pixel in UAV images. For each position of the window the  CNNs models provide  the probability of being weeds or crops. Thus, the center of the extracted image is marked by a colored dot according to the probabilities. Blue, red and white dot mean respectively that the extracted image
	is identified as weed, crop and uncertain decision (Figs. \ref{fig:spinachW-a} and \ref{fig:beanW-a}). Uncertain decision means the both probabilities are very close to 0.5. Thereafter, we used crop lines information and superpixels that have been created before, to classify all the pixels of the image. On each superpixel we look which color of dot has the majority. A superpixel is classed as crop or  weed  if the most represented dots  are  in blue respectively in red. For superpixels that have white dots as majority we used  crop lines information.  Hence, superpixels which are in the crop lines  are  regarded as crop and the others are weeds. The superpixels  created in the background are removed.  Figs. \ref{fig:spinachW-a} and \ref{fig:beanW-a} present the classification result in  parts of spinach  and bean fields. On those figures we remark that interline weeds and intra-line weeds have been detected with a low overdetection. Overdetections are mainly found on the edges of the crop rows where the window cannot overlap the whole plant. Some pixels of weeds are not entirely in red, because after performing  the threshold on the ExG, the parts of these plants which are less green are considered as soil.

	\begin{figure*}[h]
		\centering  
		\subfloat[]{\label{fig:spinachW-a}\includegraphics[height=5cm, width=7cm]{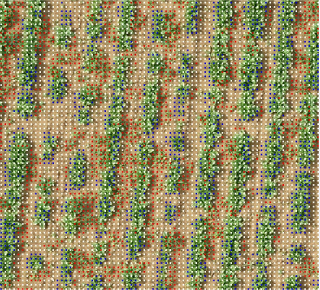}}
		\hspace{1pt}
		\subfloat[]{\label{fig:spinachW-b}\includegraphics[height=5cm, width=7cm]{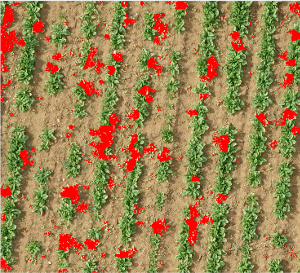}}
		\hspace{1pt}
		\subfloat[]{\label{fig:beanW-a}\includegraphics[height=5cm, width=7cm]{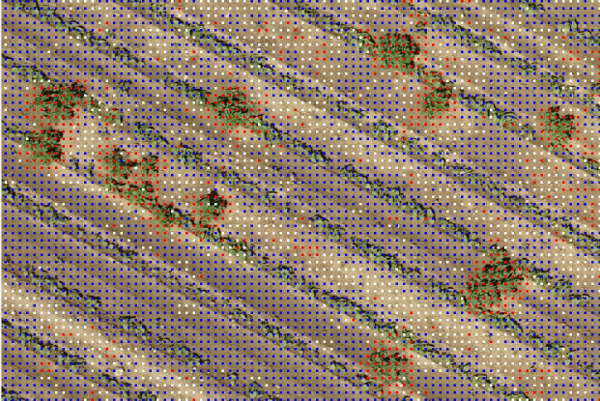}}
		\hspace{1pt}
		\subfloat[]{\label{fig:beanW-b}\includegraphics[height=5cm, width=7cm]{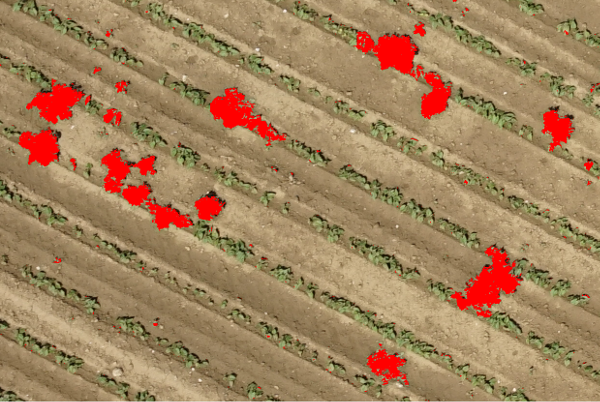}}
		\caption{Example of UAV image classification with models created by unsupervised data in two different fields.  From first  to the second line we have samples from spinach and bean fields.  On the first column we have the  samples obtained after using sliding window, without  crop lines  and background information. Blue, red and white dot mean  that the plants are is identified as weed, crop and uncertain decision respectively. On the second column  we have the detected  weeds  in red after  crop lines and background information have been applied. }
		\label{fig:classif_image}
	\end{figure*}

	However, the unsupervised data collection method  depends strongly on the efficiency of the crop line detection method and also the presence of weeds in the interline. The used  line detection  approach  has already shown its effectiveness in beet and corn fields in our previous work \cite{Bah2017}. 
	With the bean field we found even if a field does not have a lot of samples of weeds in the interline it is possible to create a robust model with data-augmentation. We also noticed using a deep learning architecture such as ResNet18 we can create  robust models for classification of weeds in bean  or spinach  fields with supervised or unsupervised data annotation. The main advantage of  our  method is  that it is fully automatic and well suited for large scale training data.

	\section{Conclusion}
	
	In this paper, we proposed a novel fully automatic learning method using Convolutional Neuronal Networks CNNs) with unsupervised training dataset collection for weeds detection from UAV images taken in bean and spinach fields.
	The results obtained have shown close performance to the supervised data labeling ones. 
	The Area Under Curve (AUC) differences are 1.5\% in the spinach field and 6\% in the bean field.  Supervised labeling is an expensive task for  human experts and according to the gaps of accuracies between the supervised and the  unsupervised labeling, our method can be a better choice in the detection of weeds, especially when the crop rows are spaced.  The proposed method is  interesting in terms of flexibility and adaptivity, since the models can be easily trained in new dataset. We  also found that ResNet18 architecture can  extract useful features for classification of weeds in bean or spinach fields with supervised or unsupervised data collection. In addition the developed method could be a key of online weeds detection with UAV.
	As future work we plan to use multispectral images because in some conditions near infra red could help to distinguish plant even if they have a similarity in the visible spectral and leave shape. With the near infra-red we plan also to improve the background segmentation.

	\section{Acknowledgment}
	This work is part of the ADVENTICES project supported by the Centre-Val de Loire Region (France), grant number ADVENTICES 16032PR. We would like to thank  the Centre-Val de Loire Region for supporting the work.
	
	

\bibliographystyle{IEEEtran}
\bibliography{mybib}

\end{document}